# How to choose "Good" Samples for Text Data Augmentation

Xiaotian Lin[1], Nankai Lin[2], Yingwen Fu[1], Ziyu Yang[1] and Shengyi Jiang[1,3](✉)

[1] School of Information Science and Technology, Guangdong University of Foreign Studies, Guangzhou, Guangdong, China
[2] School of Computer Science and Technology, Guangdong University of Technology, Guangzhou, Guangdong, China
[3] Guangzhou Key Laboratory of Multilingual Intelligent Processing, Guangdong University of Foreign Studies, Guangzhou, Guangdong, China
`jiangshengyi@163.com`

**Abstract.** Deep learning-based text classification models need abundant labeled data to obtain competitive performance. Unfortunately, annotating large-size corpus is time-consuming and laborious. To tackle this, multiple researches try to use data augmentation to expand the corpus size. However, data augmentation may potentially produce some noisy augmented samples. There are currently no works exploring sample selection for augmented samples in nature language processing field. In this paper, we propose a novel self-training selection framework with two selectors to select the high-quality samples from data augmentation. Specifically, we firstly use an entropy-based strategy and the model prediction to select augmented samples. Considering some samples with high quality at the above step may be wrongly filtered, we propose to recall them from two perspectives of word overlap and semantic similarity. Experimental results show the effectiveness and simplicity of our framework.

## 1 Introduction

Text classification is fundamental to natural language processing (NLP), which has been widely used in many fields, such as sentiment analysis and question answering. Generally speaking, a large amount of labeled data is often required to train a classifier with a strong generalization ability. Nevertheless, annotating a large-size corpus is time-consuming and laborious. To address this problem, multiple data augmentation strategies are proposed to increase the data amount for improving the model generalization ability.

Unfortunately, adequate data size does not mean good model performance. Data augmentation strategies may potentially destroy the semantic information of the text and introduce noisy samples. These noisy samples tend to negatively affect the model. Therefore, it is crucial to select the samples with high quality generated by data augmentation. However, to the best of our knowledge, there are currently no works

Xiaotian Lin, Nankai Lin and Yingwen Fu are the co-first authors. They have worked together and contributed equally to the paper.

focusing on sample selection for data augmentation in the field of NLP. Although some methods of sample selection in other domains are proposed, they focus on removing noisy samples with confident-based strategies without considering the teacher model uncertainty.

To tackle the aforementioned issue, we propose a novel sample selection framework based on self-training in this paper. The proposed framework contains two selection steps with various selection strategies to screen out the high-quality samples from data augmentation. Specifically, we firstly use an entropy-based strategy and the model prediction to select augmented samples. Considering some samples with high quality at the above step may be wrongly filtered, we propose to re-recall them from two perspectives of lexical and semantic similarity.

In summary, the main contributions are as follows.

(1) To the best of our knowledge, we are the first to focus on the sample selection problem for data augmentation in the field of natural language processing.

(2) We propose a novel sample selection framework, which can select samples from multiple dimensions.

(3) Our model can be trained automatically with little hyperparameter tuning work, which does not require normalization and probabilistic modeling.

(4) Experimental results indicate the effectiveness of our framework. Besides, we deeply analyze the reasons that affect the quality of the data augmented samples.

## 2    Related Work

### 2.1    Text Data Augmentation

Data augmentation is a highly active area of research to tackle the data-hungry problem [1]. For example, Ma et al. [2] propose to use the back-translation method to expand three Chinese datasets used for text classification, and then train and predict the datasets through a deep classification model. Tamming [3] implements and tests three methods for data augmentation: synonym replacement, back translation, and contextual augmentation. In addition, Wei and Zou [4] present a simple yet effective data augmentation technique called EDA to boost the classification performance, including synonym replacement, random insertion, random swap, and random deletion operations. Based on the EDA technique, Karimi et al. [1] propose AEDA (An Easier Data Augmentation) technique to further improve the classification performance, which includes only the insertion of various punctuation marks into the input sequence.

Pre-trained language models (PLMs) have been demonstrated that it can provide significant gains across different NLP tasks. Therefore, some researchers also have opted for using pre-trained language models for data augmentation. Based on transformer [5], Kumar et al. [6] study different types of PLMs and prove that prepending the class labels to text sequences provides a simple yet effective way to condition the PLMs for data augmentation. Hu et al. [7] utilize reinforcement learning with a conditional language model which is carried out by attaching the correct label to the input sequence when training.

Moreover, in the context of deep learning models, more and more scholars leverage the adversarial learning method to generate augmented samples. Li et al. [8]

propose a high-quality and effective method (BERT-Attack) to generate adversarial samples using pre-trained masked language models exemplified by BERT [9]. Yang et al. [10] propose a novel generative data augmentation technique to achieve accurate and robust learning in a low-resource setting. Specifically, they generate synthetic examples using PLMs and select the most informative and diverse augmented samples.

### 2.2 Sample Selection

Although data augmentation alleviates the data-hungry problem for model training to a certain extent, low-quality augmented samples would potentially bring a negative impact on the model performance. Therefore, it is particularly important to filter the noisy augmented samples. So far, there are many researches targeting sample selection.

Cao et al. [11] propose to use the uncertainty-aware self-training framework (UAST) to quantify the model uncertainty for selecting pseudo-labeling samples. Rizos et al. [12] propose an approach to improve self-training by incorporating uncertainty estimates of the underlying neural network leveraging. They leverage Monte Carlo (MC) dropout to select samples from the unlabeled pool. To address the label scarcity challenge for neural sequence labeling models, Wang et al. [13] develop self-training and meta-learning techniques to train neural sequence taggers with few labels.

There are currently no works exploring sample selection for augmented samples in nature language processing field. Different from the existing work of sample selection, we focus more on the combination of sample selection and data augmentation strategies.

## 3 Framework

As shown in Figure 1, our framework is composed of four modules: (1) Self-training module; (2) Encoder module; (3) Sample selection module; (4) Sample recall module.

We use RoBERTa as the encoder to obtain the text representations. In addition, we leverage MC dropout to compute the model uncertainty and employ the entropy-based strategy to select samples that the model is uncertain about for self-training according to this model uncertainty estimate. Finally, in the sample recall module, we automatically recall some high-quality samples with low confidence from two perspectives of word overlap and semantic fluency.

### 3.1 Self-training Module

Self-training [14] is represented as one of the semi-supervised learning methods by expanding the labeled training set to generate pseudo-labeling for unlabeled data, it is now widely used in the field of sample selection.

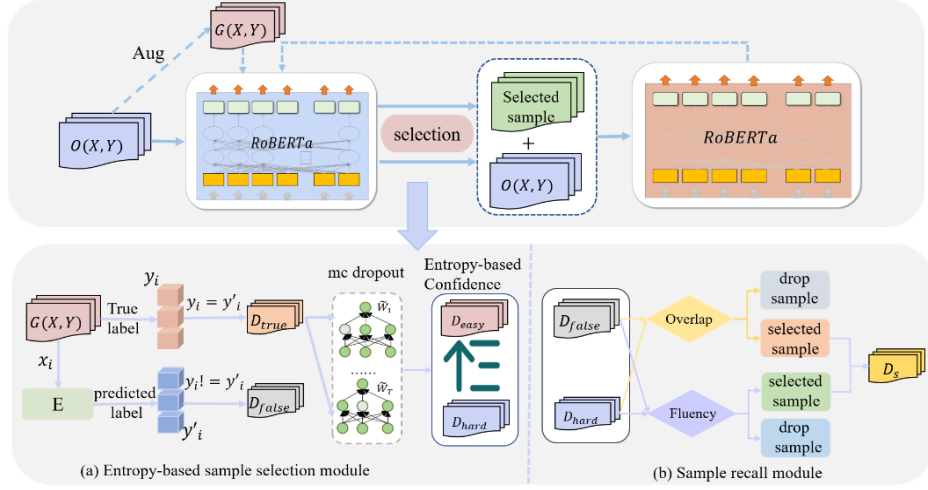

**Fig. 1.** The proposed framework structure.

Therefore, we utilize self-training technology to predict and filter the samples generated by data augmentation. Formally, given an original corpus with $N$ samples $D_l = \{x_l^i, y_l^i\}_{i=1}^{N}$ and an augmented corpus with $M$ samples $D_g = \{x_g^i, y_g^i\}_{i=1}^{M}$, the module is implemented as follows:

- Step1: Train a teacher model $RoBERTa_T$ with cross-entropy (CE) loss on the original corpus $D_l$.
- Step 2: Use the teacher network to predict the pseudo-labels $D_p$ for the augmented corpus $D_g$.
- Step 3: Select the augmented corpus $D_g$ with a variety of selection strategies to get a subset of augmented corpus $S_g$.
- Step 4: Train a student model $RoBERTa_{Stu}$ together with $S_g$ and $D_l$.
- Step 5: Treat the current student model as the teacher model, go back to step 2, and repeat steps 2 to 4 until the model converges.

### 3.2 Encoder Module

RoBERTa (Robustly Optimized BERT Pretraining) [15], a variant of BERT, only retains the masked language modeling (MLM) task for pre-training. The first and last positions in the input sentence would be respectively given special tokens [CLS] and [SEP]. For each token in an input sentence, its input representation is constructed by summing up its corresponding token, segment, and position embeddings.

We leverage RoBERTa as the encoder to train the teacher and student models in this paper. For the classification task, the final hidden state of [CLS] is usually utilized to present the sentence. Hence, for the $i$-th sentence, the sentence representation is calculated as follows:

$$S_i = RoBERTa(a_i, b_i, c_i)$$

where $a_i, b_i, c_i$ are the token, segmentation, and position embeddings.

### 3.3 Sample Selection Module

Since some noisy samples are generated by data augmentation, the model tends to gradually bias towards the noisy data during self-training, affecting the model performance. Previous works have focused on using the confident-based method to remove noisy samples without considering the uncertainty of the teacher model. Inspired by these methods, we propose a two-stage sample selection method to solve this problem, which includes clean data selection and confidence evaluation.

**Clean Data Selection.** To verify the correctness of the predicted labels, a binary classification task is introduced, which aims to filter out the augmented data with wrongly predicted labels. We use the teacher model to predict the labels of the augmented samples and compare the predicted labels with their ground truths. Formally, given an augmented sample $x_i^g$ with its ground truth $y_i^g$, we first use the encoder with a linear classify to predict its probability distribution $f(x_i^g, \theta)$ and obtain the pseudo-label $\tilde{y}_i^g$.

$$f(x_i^g, \theta) = softmax(W x_i^g + b)$$
$$\tilde{y}_i^g = argmax\, f(x_i^g, \theta)$$

Then, we regard the samples with the same predictions and ground truths as clean samples and put them into the set $D_{clean}$ while others are put into the set $D_{noisy}$.

**Confidence Evaluation.** After filtering the wrongly predicted samples, based on the probability distribution, we utilize entropy-based confidence measurement to quantify the confidence of the samples on $D_{clean}$. To better evaluate the sample confidence, we first leverage the MC dropout that conducts $T$ forward passes with dropout layers to predict labels for augmented samples. For example, given an augmented sample $x_i$, its probability of class $c$ in the $t$-th dropout $p_{ic}^t$ is as follows:

$$p_{ic}^t = p(y = c|x_i) = softmax(\widetilde{W}_t x_i + B)$$

Then, according to the probability distribution $p_{ic}^t$, we aggregate predictions from $T$ passes and utilize entropy-based confidence measurement to generate sample confidence which is calculated as follows:

$$H(p_i) = -\frac{1}{C}\sum_{c=1}^{C}\frac{1}{T}\sum_{t=1}^{T} p_{ic}^t \log(p_{ic}^t)$$
$$w_{easy}^i = 1 - H(p_i)$$
$$w_{hard}^i = H(p_i)$$

where $w_{easy}^i$ and $w_{hard}^i$ respectively donate the easy and hard confidences for the $i$-th samples. Eventually, we rank all the confidence of the samples and get two sets $D_{easy}$ and $D_{hard}$.

### 3.4 Sample Recall Module

Some high-quality samples in $D_{noisy}$ and $D_{hard}$ would be wrongly filtered out, which can improve the model performance to a certain extent. Therefore, we propose to re-recall them from two perspectives of word overlap and semantic fluency.

**Word Overlap.** Word overlap measures the text similarity from the perspective of words. It can effectively filter the samples which are similar in meaning to the original sentence. In this paper, we use the Jaccard coefficient to calculate the word overlap between the augmented samples and the original samples. Suppose $x_l^i$ and $x_g^i$ represent the original and the augmented samples in $i$-th sample, the word overlap $J(x_l^i, x_g^i)$ can be calculated as follows:

$$J(x) = J(x_l^i, x_g^i) = \frac{|x_l^i \cap x_g^i|}{|x_l^i \cup x_g^i|} = \frac{|x_l^i \cap x_g^i|}{|x_l^i| + |x_g^i| - |x_l^i \cap x_g^i|}$$

**Semantic Fluency.** Fluency can be captured by statistical language models [16], which is important for grammatical error correction task [17][18]. Therefore, we propose a metric to evaluate the semantic fluency of the augmented data. Formally, given an original sample $x_l^i = \{w_1^i, w_2^i, \ldots, w_n^i\}$ with $n$ words and its augmented sample $x_g^i = \{u_1^i, u_2^i, \ldots, u_n^i\}$ with $m$ words, we first compute their perplexity $P(\cdot)$ using the language model ResLSTM [19] trained on the One-Billion[1] corpus.

$$P(x_l^i) = \sqrt[n]{\prod_{j=1}^{n} \frac{1}{p(w_j^i | w_1^i w_2^i \ldots w_{j-1}^i)}}$$

$$P(x_g^i) = \sqrt[m]{\prod_{j=1}^{m} \frac{1}{p(u_j^i | u_1^i u_2^i \ldots u_{j-1}^i)}}$$

The semantic fluency $F_i$ for the $i$-th sentence is then calculated based on the perplexity.

$$F(x) = F(x_l^i, x_g^i) = P(x_l^i) - P(x_g^i)$$

After that, to recall the high-quality samples from $D_{noisy}$ and $D_{hard}$, we use $D_{easy}$ to automatically calculate the thresholds to alleviate the workload of normalization and probabilistic modeling. Specifically, we calculate the average Jaccard coefficient $J_{avg}$ and fluency $F_{avg}$ of the samples on $D_{easy}$ and use them as the thresholds to select the recalled samples.

$$R_{cutoff} = \{x \in D_{noisy} + D_{hard} | J(x) > J_{avg} \land F(x) < F_{avg}\}$$

## 4 Experiment

### 4.1 Dataset

Stanford Sentiment Treebank (SST) [20] is one of the most popular sentiment classification datasets, which is composed of three parts: SST-1, SST-2, and SST-5. SST-1 and SST-2 are binary classification datasets and SST-5 contains five sentiment labels for fine-grained sentiment classification task. In this paper, we conduct extensive experiments on SST-5. As shown in Table 1, the original training, validation, and test set of this dataset contain 8543, 1100, and 2209 samples respectively. We directly use them to train and test the models. In addition, we expand the training data using two data augmentation methods. (1) **Easy Data Augmentation (EDA)** [4]. This augmentation method includes synonym replacement (SR), random insertion (RI), random swap (RS), and random deletion (RD). Each of them doubles the amount of the original training set. Finally, we construct a new training set that is 4 times larger than the original training set. (2) **nlpaug tool**[2]. We randomly select four augmentation methods from this tool to augment each original sample. This can generate another new training set that is 4 times larger than the original training set.

---

[1] https://github.com/ciprian-chelba/1-billion-word-language-modeling-benchmark

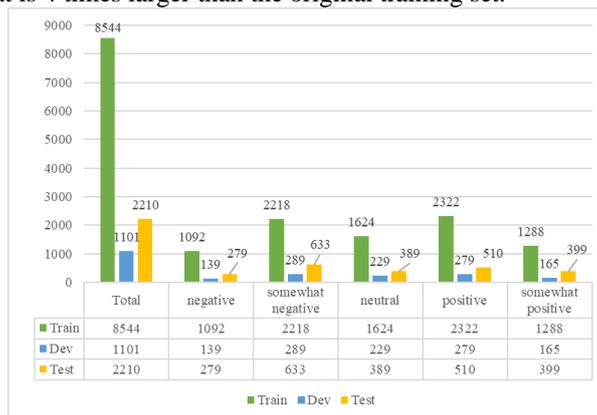

**Fig. 2.** The dataset distribution.

### 4.2 Experiment Settings

We implement our models in Pytorch[3] and transformers[4] frameworks. In the training process, we use Adam as the optimizer and report the results over 3 runs with different random seeds. We compare the performance of different PLMs including, RoBERTa-base[5], BERT-base[6], XLM-RoBERTa[7], and mBERT[8], to select the best base model. Eventually, we choose RoBERTa-base as our base model. The hyperparameters of the proposed framework and baselines are shown in Table 1.

As for the evaluation metric, accuracy implemented by scikit-learn[9] is utilized for model evaluation as in previous works [21][22][23][24][25][26][27][28][29].

**Table 1.** The value of the parameters.

| Module | Parameter | Value |
| --- | --- | --- |
| ResLSTM | Embedding size | 256 |
|  | Token length | 256 |
|  | Batch size | 32 |

---

[2] https://github.com/makcedward/nlpaug
[3] https://github.com/pytorch/pytorch
[4] https://github.com/huggingface/transformers
[5] https://huggingface.co/roberta-base
[6] https://huggingface.co/bert-base-uncased
[7] https://huggingface.co/xlm-roberta-base
[8] https://huggingface.co/bert-base-multilingual-uncased
[9] https://scikit-learn.org/stable/

| | | |
|---|---|---|
| | Epoch | 20 |
| | Learn rate | 0.1 |
| | LSTM Layer | 2 |
| Encoder Moule | Max length | 32 |
| | Learning rate | 5e-5 |
| | Batch size | 32 |
| | Epoch | 15 |
| | Dropout | 0.5 |
| | Weight decay | 0.001 |
| | Optimization function | Adam |
| Sample selection module | Entropy threshold with EDA | 0.2 |
| | Entropy threshold with nlpaug | 0.25 |
| | $T$ | 10 |

## 5 Result and Analysis

### 5.1 Main Performance

To verify the effectiveness of our framework, we compare our model with multiple advanced baseline methods. The results are shown in Table 2. Our model achieves the competitive results with the accuracy value reaching 56.6 in EDA and 56.0 in nlpaug compared to the models with the same base model. These results demonstrate the effectiveness of our selection strategies.

**Table 2.** Main performance.

| Model | Accuracy |
|---|---|
| BERT Large [21] | 55.5 |
| BCN+ELMo [22] | 54.7 |
| byte mLSTM7 [23] | 54.6 |
| BCN+Char+CoVe [21] | 53.7 |
| BERT Base [24] | 53.2 |
| Star-Transformer [25] | 53.0 |
| RoBERTa-Base | 54.2 |
| SELFEXPLAIN [26] | 54.3 |
| MS-Transformer [27] | 53.5 |
| Sentix [28] | 55.6 |
| ACNN-BERT [29] | 53.7 |
| Our method with EDA | **56.5** |
| Our method with nlpaug | 56.0 |

### 5.2 Effectiveness of Different PLMs

Table 3. Effectiveness of different PLMs.

| Model | Accuracy |
|---|---|
| RoBERTa-Base | 54.2 |
| BERT-base | 53.1 |
| mBERT | 48.4 |
| XLM-RoBERTa | 51.8 |

The results regarding different PLMs as the encoder are shown in Table 3. It is obvious that the RoBERTa-Base model achieves the best performance and we finally choose it as our base model.

### 5.3 Effectiveness of Different Augmentation Methods

To explore the effectiveness of different data augmentation methods, we experiment on each data augmentation method and compare them with the models trained with all augmentation data methods or without any augmentation data. As shown in Table 4, it seems that the EDA(SR)[10] and EDA(RS) strategies have a slight improvement in the model performance. The EDA(SR) strategy does not change the text's basic semantic information. Additionally, the EDA(RS) strategy only swaps the token order, and this change does not affect the model performance too much [30]. On the contrary, the EDA(RI) and EDA(RD) strategies impair the model performance, and we hypothesize the reason is that such strategies tend to change the text semantic information. In addition, we can see that training the model with the full augmented data does not bring significant improvement, which also hints at the effectiveness of our framework.

Table 4. Effectiveness of different augmentation methods.

| Model | Accuracy |
|---|---|
| RoBERTa-Base | 54.2 |
| RoBERTa-Base with EDA (full data) | 54.4 |
| RoBERTa-Base with nlpaug (full data) | 53.6 |
| RoBERTa-Base with EDA(SR) | 54.8 |
| RoBERTa-Base with EDA(RI) | 53.6 |
| RoBERTa-Base with EDA(RS) | 54.6 |
| RoBERTa-Base with EDA(RD) | 52.9 |

### 5.4 Effectiveness of Each Module

---

[10] EDA(·) means the EDA method using only one strategy.

**Table 5.** Effectiveness of each module.

| Model | Num. of augmented data | Acc |
|---|---|---|
| BERT Base | - | 54.2 |
| BERT Base with EDA | 34176 | 54.4 |
| + Entropy-based sample selection module | 584 | 55.5 |
| + Recall module | 23765 | 56.6 |

In addition, we verify the effectiveness of each module in our framework. As the result in Table 5, each module contributes to the model performance. Especially, compared with the RoBERTa-Base model, the model with the sample selection module is increased by 1.3% in accuracy. Since the reliability module filters a large number of noisy samples that are harmful to the model performance, it can improve the model performance to a certain extent. Besides, the recall module can also bring a certain improvement to the model performance, which demonstrates the effectiveness and importance of the recall module. Since the recall module can recall some high-quality samples from $D_{noisy}$ and $D_{hard}$ for further improvement.

### 5.5 Effectiveness of Different Thresholds

We explore the best combination of thresholds in different strategies. Specifically, we sequentially explore the optimal thresholds for different strategies in different modules. As shown in Figure 3, the entropy strategy performs best when the threshold is 0.25 in EDA and 0.3 in nlpaug.

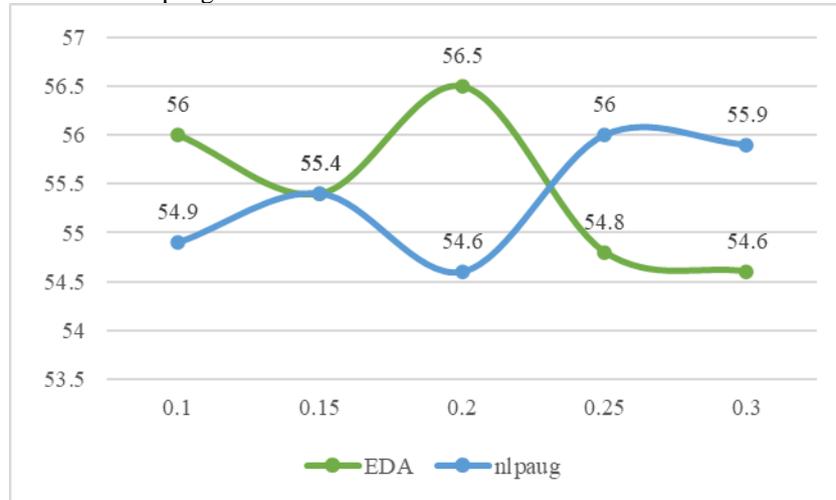

**Fig. 3.** The structure of our framework.

**Table 6.** The augmented data number.

| Strategy | Num. of augmented data |
|---|---|
| Synonyms Replace (SR) | 5227 |
| Randomly Insert (RI) | 6093 |
| Randomly Swap (RS) | 6265 |
| Randomly Delete (RD) | 6180 |
| Total | 23765 |

To explore the impact of different data augmentation methods on model performance, we take the results of EDA as an example and count the number of final retained samples under each operation to determine which operation was the least able to generate enhanced samples that were beneficial to the model. Eventually, after the joint selection of the two modules, we obtain 23,765 high-confidence samples in total, of which the samples using SR strategies had the lowest recall rate with only 5,227 recalled samples. The number of the recalled augmented data is shown in Table 6.

## 6   Conclusion

In this paper, we propose a sample selection framework for the data augmentation technique, which is easy and reasonable to utilize. Aiming at the existing problems of data augmentation, the proposed framework introduces various selection strategies to select the high-quality samples from the augmented samples, which can effectively enhance the model performance. In the future, we would keep on improving the model performance and applying our framework to other domains.

## 7   Declarations

### 7.1   Conflicts of interests

The authors declare that we do not have any commercial or associative interest that represents a conflict of interest in connection with the work submitted and that the research do not involve human participants and/or animals.

### 7.2   Data availability

Data will be made available on request.